\documentclass[sigconf]{acmart}

\usepackage{multirow}
\usepackage{subfig}
\usepackage{graphicx}
\usepackage{algorithmic}
\usepackage{algorithm}
\usepackage{colortbl}
\usepackage{makecell}

\newlength\savewidth

\AtBeginDocument{%
  }

\setcopyright{acmlicensed}
\copyrightyear{2018}
\acmYear{2018}
\acmDOI{XXXXXXX.XXXXXXX}

\acmConference[Conference acronym 'XX]{Make sure to enter the correct
  conference title from your rights confirmation emai}{June 03--05,
  2018}{Woodstock, NY}
\acmISBN{978-1-4503-XXXX-X/18/06}

\begin{document}

\title{EntityCLIP: Entity-Centric Image-Text Matching via Multimodal Attentive Contrastive Learning
}


\author{Yaxiong Wang}
\email{wangyx15@stu.xjtu.edu.cn}
\affiliation{%
  \institution{Hefei University of Technology}
  \streetaddress{No. 193 Tunxi Road}
  \city{Hefei}
  \state{Anhui}
  \country{P.R. China}
  }
  
\author{Yujiao Wu}
\email{yujiaowu111@gmail.com
}
\affiliation{%
  \institution{CSRIO}
  \city{Canberra}
  \country{Australia}
  }

\author{Lianwei Wu}
\email{wlw@nwpu.edu.cn}
\affiliation{%
  \institution{Northwestern Polytechnical University}
  \streetaddress{No. 1 Dongxiang Road}
  \city{Xi'an}
  \state{Sha'xi Province}
  \country{P.R. China}
  }

\author{Lechao Cheng}
\email{chenglc@hfut.edu.cn}
\affiliation{%
  \institution{Hefei University of Technology}
  \streetaddress{No. 193 Tunxi Road}
  \city{Hefei}
  \state{Anhui}
  \country{P.R. China}
  }

\author{Zhun Zhong}
\email{zhunzhong007@gmail.com}
\affiliation{%
  \institution{Hefei University of Technology}
  \streetaddress{No. 193 Tunxi Road}
  \city{Hefei}
  \state{Anhui}
  \country{P.R. China}
  }

\author{Meng Wang}
\email{eric.mengwang@gmail.com}
\affiliation{%
  \institution{Hefei University of Technology}
  \streetaddress{No. 193 Tunxi Road}
  \city{Hefei}
  \state{Anhui}
  \country{P.R. China}
  }

\renewcommand{\shortauthors}{Yaxiong Wang et al.}

\begin{abstract}
Recent advancements in image-text matching have been notable, yet prevailing models predominantly cater to broad queries and struggle with accommodating fine-grained query intention.
In this paper, we work towards the \textbf{E}ntity-centric  \textbf{I}mage-\textbf{T}ext \textbf{M}atching (EITM), a task that the text and image involve specific entity-related information. The challenge of this task mainly lies in the larger semantic gap in entity association modeling, comparing with the general image-text matching problem.
To narrow the huge semantic gap between the entity-centric text and the images, we take the fundamental  CLIP as the backbone and devise a multimodal attentive contrastive learning framework to tam CLIP to adapt EITM problem, developing a model named EntityCLIP.
The key of our multimodal attentive contrastive learning is to generate interpretive explanation text using Large Language Models (LLMs) as the bridge clues.
In specific, we proceed by extracting explanatory text from off-the-shelf LLMs. This explanation text, coupled with the image and  text, is then input into our specially crafted Multimodal Attentive Experts (MMAE) module, 
which effectively integrates explanation texts to narrow the gap of the entity-related text and image in a shared semantic space. 
Building on the enriched features derived from MMAE, we further design an effective Gated Integrative Image-text Matching (GI-ITM) strategy. The GI-ITM employs an adaptive gating mechanism to aggregate MMAE's features, subsequently applying image-text matching constraints to steer the alignment between the text and the image.
Extensive experiments are conducted on three social media news benchmarks including N24News, VisualNews, and GoodNews, the results shows that our method surpasses the competition methods with a clear margin. 
\end{abstract}


\maketitle

\begin{figure}
    \centering
    \includegraphics[width=1\linewidth]{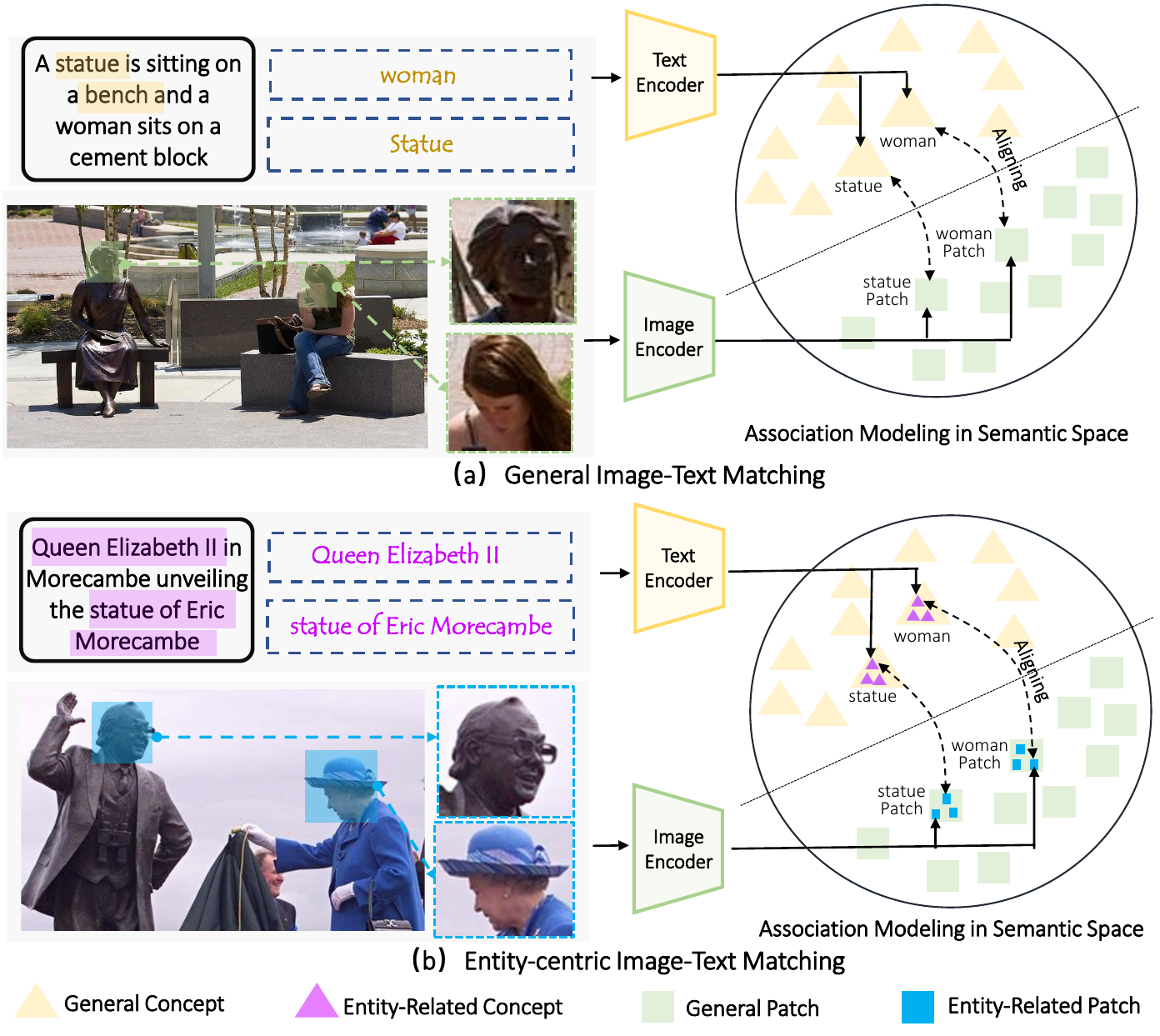}
    \caption{In comparison with general image-text matching (subfigure (a)), Entity-centric Image-text matching (EITM) requires the model to learn deeper by understanding and discriminating the specific entities under the general concepts (subfigure (b)). For example,``Queen Elizabeth II" in woman, and ``statue of Eric Morecambe" in statue. This specificity introduces a substantial semantic gap, presenting a significant challenge for cross-modal retrieval.}
    \label{teaser}
\end{figure}

\section{Introduction}
Image-text matching is a fundamental cross-modal task and has a long line of research~\cite{vse++,dpc,scan,pfan,pfan++}. The key of these endeavors lies in the development of a unified semantic space that facilitates the quantification of the relationship between textual and visual content, thereby enabling the determination of their relative rankings. However, the prevailing approaches in image-text matching predominantly concentrate on abstract, high-level semantic features. As depicted Figure~\ref{teaser} (a), queries are often formulated with broad semantic terms such as "woman" or "statue". While this use of general descriptors simplifies the retrieval process by allowing models to understand general concepts more readily, it falls short in addressing the finer-grained queries, for example entity-centric image retrieval. 

EDIS~\cite{edis} is the pioneering work to emphasize the significance of entity-centric information retrieval (ECIR) in practical scenarios, and the authors propose a benchmark for multimodal news (image \& headline) retrieval with entity text as the query. In contrast, this paper considers a more general and challenging ECIR problem: entity-centric image-text matching (EITM). Particularly, EITM is EDIS eliminating the entity text (\emph{i.e.} headline) associated with the image, as a result, EITM is a more challenging problem since the model needs to directly estimate the association between the query text and the image without any auxiliary information. What's more, not all images come with accompanying headlines as considered in EDIS, which is another reason for exploring EITM task.

\begin{figure*}
    \centering
    \includegraphics[width=0.95\linewidth, height=0.55\linewidth]{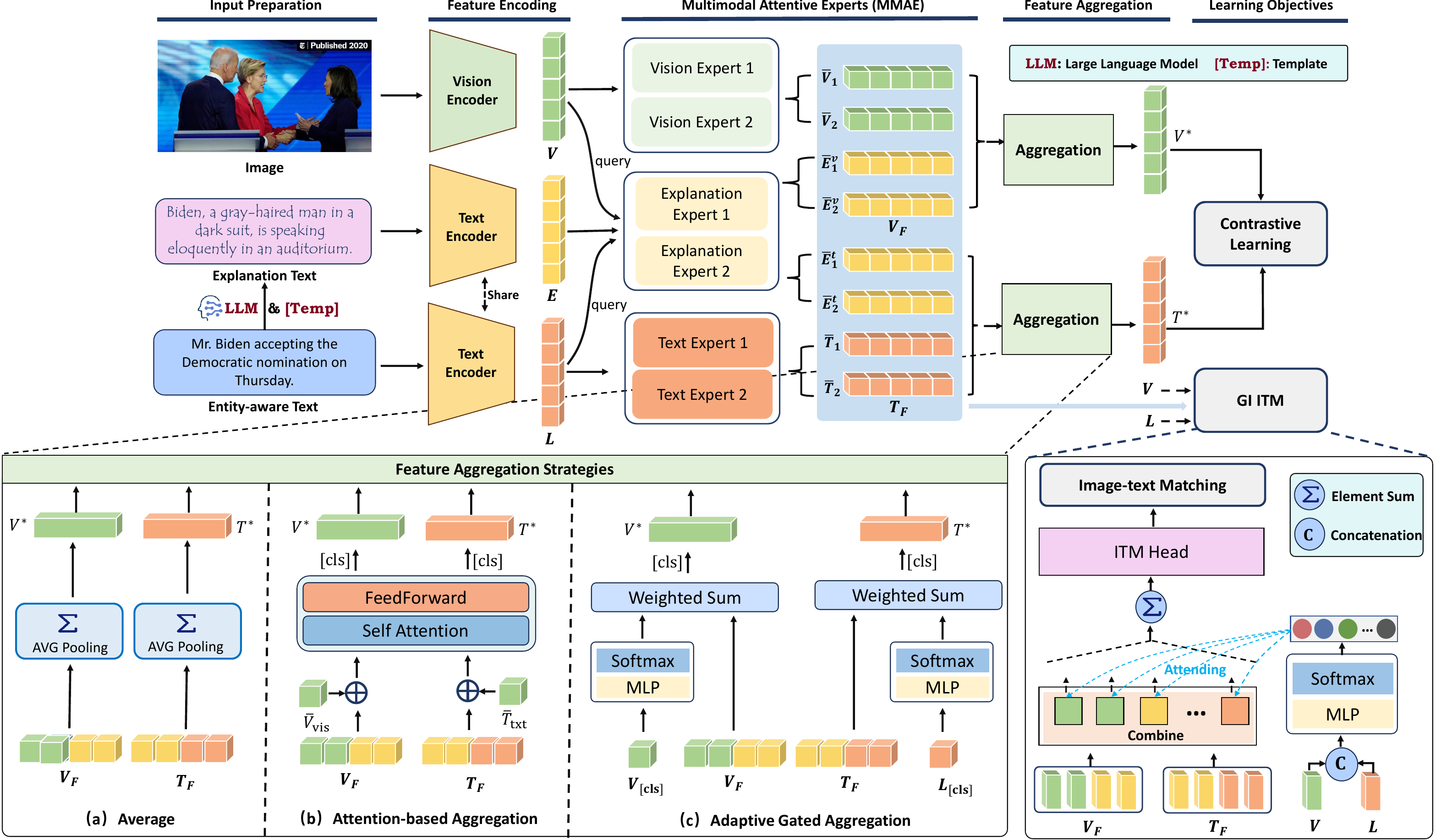}
    \caption{Illustration of training EntityCLIP. Initially, we harness Large Language Models (LLMs) to generate explanation text based on the entity-text query. This text, along with the query and image, is then encoded to derive representations. These are subsequently processed by the Multimodal Attentive Experts (MMAE) to integrate the query and image features, leveraging the explanation text to bridge semantic disparities. The framework is optimized through contrastive learning, coupled with a Gated Image-text Matching loss to refine the alignment and learning of the network.
    }
    \label{framework}
\end{figure*}

In contrast to conventional image-text matching, the textual query in EITM is characterized by its specificity, as illustrated in Figure~\ref{teaser} (b). For instance, queries such as ``Queen Elizabeth II" and ``statue of Eric Morecambe" are employed to pinpoint exact images. However, this precision introduces a significant challenge: the need for robust entity understanding and the accurate association of these entities with image content. This presents a semantic gap that is more pronounced than in generic image-text matching scenarios. 
For the query "Queen Elizabeth II in Morecambe unveiling the statue of Eric Morecambe", 
it is extremely difficult for the model to find the expected images without knowing the visual appearance of "Queen Elizabeth II" and ``statue of Eric Morecambe".
A possible solution to understand the entities is to query the external library knowledge base like  Wikipedia. However, most external knowledge bases usually introduce the entity in an extremely detailed fashion and contains many information that is not related to the expected visual contents. Besides, the incorporation of such databases can substantially increase the complexity and computational overhead of the framework.

Advancements in large language models (LLMs) and multimodal foundation models (MFMs)  offer us another chance to well understand the entities.
LLMs, such as ChatGPT, LLAMA~\cite{llama}, and Mistral~\cite{mistral}, with their expansive parameter sets and training on extensive, varied datasets, embody a rich repository of real-world knowledge. These models can be effectively utilized to extract meta information pertaining to specific entities, thereby serving as a valuable tool for querying entity-related details. The obtained meta information of the entities offers explicit and effective insights to bridge the semantic gap inherent in EITM. Concurrently, MFMs, particularly those designed for retrieval tasks like CLIP, facilitate the alignment of visual and textual data through the use of vast image-text pairs. This alignment provides an advantageous initialization for entity representation, supplying implicit contextual clues that augment the understanding of entities within EITM problem. The synergistic application of LLMs for explicit bridging clues and MFMs for implicit entity representation presents a promising avenue for addressing the semantic gap in EITM.


With the above considerations, we take LLMs as the external knowledge base and the CLIP as the backbone network to address the problem of EITM. We meticulously craft prompts to elicit entity-specific explanations from LLMs. Utilizing CLIP encoders, we extract representations for the image, query text, and explanation text. Subsequently, a Multimodal Attentive Experts (MMAE) module is designed to harness the explanation text effectively. Within MMAE, visual and textual experts encode the respective features, while explanation experts leverage the image and text to distill insights from the explanation text, thus bridging the semantic gap between the entity-centric query and candidate images. The resultant visual and textual features from both pure and explanation experts are consolidated to form the definitive image and text features. Finally, contrastive learning is applied to optimize the network.

Our practice shows that MMAE is an effective design to narrow the semantic gap, the produced features are comprehensive representation for the input image and the query. To further utilize these informative features from MMAE, we propose a Gated Integrative Image-text Matching (GI-ITM) mechanism. Particularly, the intermedia features from MMAE are first aggregated with a gated integrative mechanism to form a multimodal representation, which is then fed through an image-text matching module to further align the cross-modal inputs. 
MMAE and GI-ITM only acts during training to optimize CLIP, yet they are not utilized during inference. As a result, EntityCLIP retains the efficiency of the original CLIP model. In summary, we highlight the contributions of this paper as follows:
\begin{itemize}
    \item We make an early exploration of the entity-centric Image-text matching, and propose EntityCLIP for this problem with the aid of the LLMs and MFMs.  
    \item A multimodal attentive experts (MMAE) is proposed to aggregate the entity-specific explanation text to bridge the huge semantic gap between the entity-centric text and the image. 
    \item A gated integrative image-text matching (GI-ITM) is designed to fuse the comprehensive features of MMAE, coupled with a image-text matching module to enhance the cross-modal alignment.
\end{itemize}


\begin{figure}
    \centering
    \includegraphics[width=0.92\linewidth]{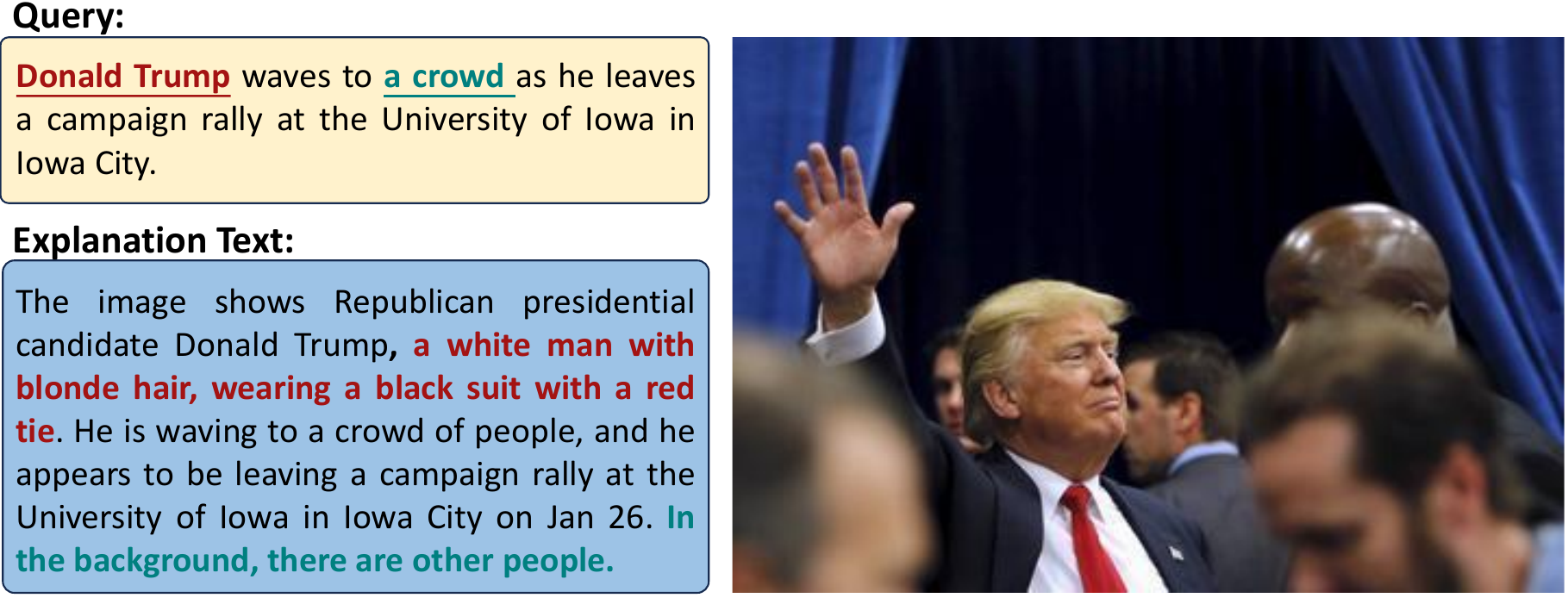}
    \caption{Explanation text example for an entity-centric query. The explanation text can offer visual details regarding the entities of \emph{Donald Trump}, and further explain some occasion like \emph{the crowd}, thereby narrowing the semantic gap.}
    \label{exptxt}
\end{figure}


\section{Related Works}
\noindent\textbf{Image-Text Matching} boasts an extensive research trajectory, with prevailing methodologies delineated into two principal learning paradigms: end-to-end training and pretraining-based approaches. End-to-end training, which encompasses diverse architectural implementations, predominantly engenders model learning specific to target datasets such as Flickr30K~\cite{flickr30k}, MS-COCO~\cite{mscoco}. Noteworthy models within this paradigm include VSE++~\cite{vse++}, SCAN~\cite{scan} \emph{et al.}
In contrast, pretraining-based models undergo a bifurcated training regimen, encompassing both a pretraining phase and a subsequent fine-tuning phase. During the pretraining phase, models are endowed with extensive image-text datasets culled from social media platforms. This foundation is followed by fine-tuning, where models are refined using samples from the target dataset, thereby striving for enhanced performance. Leveraging large-scale datasets, pretraining-based methods often substantially outperform end-to-end training methodologies.
The CLIP~\cite{clip} model stands as a prominent exemplar of multimodal pretraining, consistently achieving state-of-the-art results across a variety of tasks. Nonetheless, existing methodologies predominantly concentrate on coarse image retrieval predicated on general textual descriptions. In stark contrast, our study focuses on fine-grained, entity-oriented image retrieval—a domain that has received scant attention in prior research endeavors.

\noindent\textbf{Large Language Models} have emerged as potent and versatile tools in recent years, revolutionizing the field of natural language processing~\cite{gpt4,chatgpt,mistral}. Predecessors to models like ChatGPT, such as GPT 1-3~\cite{gpt1,gpt2,gpt3}, are primarily utilized for advancing text encoding and recognition tasks. ChatGPT represents a significant leap forward, being trained with the innovative Reinforcement Learning from Human Feedback (RLHF) technique, marking it as the first in its class of general and intelligent models. The subsequent GPT-4~\cite{gpt4} has furthered this progression, refining the capabilities of its predecessor.
The evolution continues with the development of models like the LLAMA series~\cite{llama,lamma2}, Mistral~\cite{mistral}, and ChatGLM~\cite{chatglm}, each trained on vast repositories of natural language data and employing sequence prediction to internalize a broad spectrum of knowledge. This endows LLMs with the characteristics of a knowledge base, offering profound implications for information retrieval.
In the context of this work, which is centered on entity-oriented retrieval, we advocate for harnessing LLMs to extract entity-centric information. This approach is designed to bridge the substantial semantic gap that often exists between textual queries and images, thereby enhancing the precision and relevance of retrieval outcomes.

\section{Methodology}
\textbf{Overview.} 
Figure~\ref{framework} depicts the framework of our EntityCLIP during training phase. EntityCLIP ingests an entity-centric textual query, its corresponding image, and the associated explanation text. These inputs undergo feature extraction via respective image and text encoders. The resultant features then proceed to our Multimodal Attentive Experts (MMAE) module, where the explanation text acts as a semantic bridge, aligning the image and query representations within a unified feature space. We further refine the network through contrastive learning, supplemented by an image-text loss from our Gated Integrative Image-text Matching (GI-ITM) strategy to enhance the cross-modal alignment.


\subsection{Input Preparation.} 
\label{expgene}
This stage focuses to generate the explanation text for the text-image pair. Given the query text $T$ and a large language model, 
we first  design a prompt template: 

\texttt{I have an image: [T]. Describe the image content matching this description in detail, your answer should include more appearance description of the mentioned person, object, place or occasion.}  

\noindent Subsequently, we input this template into an off-the-shelf LLM to generate the explanation text $E$.


{As shown in Figure~\ref{exptxt}}, the explanatory text derived from Large Language Models (LLMs) encapsulates visual attributes of entities that correspond to the contents observed in the images, while also maintaining the relational coherence of these entities with the query text. This alignment forms an effective intermediary, adept at narrowing the semantic gap that exists between entity-centric queries and their corresponding images.


\begin{table*}[t]
\centering
\caption{\textbf{Quantitative results on N24News}. Three comparison groups from top to bottom subsequently shows the zero-shot performance, fine-tuned performance, and the variants of our EntityCLIP. \textbf{Bold} indicates the best perforamnce. 
}
\vspace{-0.4cm}
\setlength{\tabcolsep}{2mm}
\setlength{\extrarowheight}{-4pt}
{\renewcommand\arraystretch{0.65}\begin{tabular}{@{}lccccccccccc@{}}
\toprule[1.5pt]
\rowcolor[gray]{.95}\multirow{2}{*}{Models}  &\multicolumn{3}{c}{Image Retrieval} &\multirow{2}{*}{AVG t2i$\uparrow$} &\multirow{2}{*}{MR t2i$\downarrow$} & \multicolumn{3}{c}{Text Retrieval} &\multirow{2}{*}{AVG i2t$\uparrow$} &\multirow{2}{*}{MR i2t$\downarrow$}  \\ 
\cmidrule(lr){2-4} \cmidrule(lr){7-9}
\rowcolor[gray]{.95}& $K=1$ & $K=5$ & $K=10$  & &  & $K=1$ & $K=5$ & $K=10$  & &\\

\midrule
XFM~\cite{xfm} & 7.10 & 18.17  & 24.51	 &30.08	 &420.00	 &7.87	 & 17.21	& 23.17 &29.03	 & 418.29	 \\
X-VLM~\cite{xvlm}  & 10.57  & 22.52	 & 29.01	& 32.95 &457.98	 & 12.05	 & 24.35	 &30.94 & 34.88 &372.05 	 \\
X2-VLM~\cite{x2vlm}  & 10.66 & 22.80  & 29.88 & 34.32 & 380.00 & 9.34 & 19.17 & 24.34 & 29.51 &	470.75 \\
ALBEF~\cite{albef}  & 21.18  & 38.08  & 46.15  &48.44  &234.75 	 & 21.49	 & 38.26 & 45.97  &48.81 & 198.52 \\
EDIS~\cite{edis}  & 33.70 & 53.89 &62.41 & 62.51 & 126.54 &  33.18 &	54.02 & 62.02 & 62.47 & 111.76 \\
BLIP~\cite{blip}  & 30.31 	&50.854	&58.90 & 59.83 & 133.23 &30.83	&51.72	&60.10 & 60.98 & 100.59 \\
BLIP2~\cite{blip2}  & 33.02	&53.56	&61.40 & 61.99 & 121.40 & 32.28 &	53.53 &62.18 & 62.34 & 130.84 \\
$\text{CLIP}_{\text{ViT-B/32}}$~\cite{clip}  & 48.27 &69.95 &77.11 & 75.35 & 52.087 & 43.27 &	64.42 &71.86 & 70.99 & 59.00 \\
$\text{CLIP}_{\text{ViT-B/16}}$~\cite{clip}  & 55.01 &75,74 &81.79 &79.79 & 39.83 & 49.33 &69.92 &76.16 & 75.26 & 54.98 \\
\midrule

ALBEF~\cite{albef}  & 39.80  & 51.72  & 60.35  & 58.38  & 80.91 	 & 41.42	 & 53.72 & 62.37  & 60.88 & 64.75 \\
BLIP~\cite{blip} & 44.39 	& 66.98	& 73.10 & 69.34 & 69.34 & 46.04	& 67.13	& 75.12 & 71.28 & 52.10 \\
$\text{CLIP}_{\text{ViT-B/32}}$~\cite{clip}   & 51.52	 &73.87	&81.40	 &78.84	&27.84  &50.90	&73.65	&81.09  & 78.74 & 22.55\\
$\text{CLIP}_{\text{ViT-B/16}}$~\cite{clip} & 56.50 &78.66	&84.63 &82.25 &19.54 &57.45	&78.72	&85.11 &82.68 &14.59 \\
\midrule
\textcolor{white}{$\text{EntityCLIP}_{\text{ViT-B/32}}$}\qquad AVG & 53.27 &76.44 &83.20 & 80.49 & 23.24 & 54.57	&76.39	&83.46 &81.06 & 16.06 \\     
$\text{EntityCLIP}_{\text{ViT-B/32}}$\qquad Attn &53.87 &76.57	 &\textbf{83.77} & \textbf{80.76} &23.30 & 55.26 &\textbf{76.58} &83.80 &81.29 &15.79 \\
\textcolor{white}{$\text{EntityCLIP}_{\text{ViT-B/32}}$}\qquad AGA &  \textbf{54.08} & \textbf{76.58}	&83.49 & {80.69} & \textbf{23.09} &\textbf{55.90} &76.44 &\textbf{83.94} & \textbf{81.87} & \textbf{15.48}  \\
\textcolor{white}{$\text{EntityCLIP}_{\text{ViT-B/16}}$}\qquad  AVG & 60.48	&80.04	&86.19 &83.69 &18.86 &61.58	&81.17	&86.76 &84.47 &12.19 \\     
$\text{EntityCLIP}_{\text{ViT-B/16}}$\qquad Attn & 60.79	 &\textbf{80.91}	&86.43 &84.06 &18.04 & 61.15 &81.47 &\textbf{86.94} &84.52 &11.64 \\
\textcolor{white}{$\text{EntityCLIP}_{\text{ViT-B/16}}$}\qquad AGA & \textbf{60.85} & 80.70 &\textbf{86.69} & \textbf{84.74} &\textbf{17.53} & \textbf{61.94} & \textbf{81.55}	 &{86.86} & \textbf{84.75} & \textbf{11.02} \\	
\bottomrule[1.5pt]
\end{tabular}}
\label{n4news}
\end{table*}

\subsection{Mutimodal Attentive Experts}
\textbf{Feature Encoding} stage encodes the raw data into the fundamental representations. 
Given the query text $T$, matched image $V$, and the generated explanation text $E$, we first feed the image and texts into the transformer-based vision encoder and text encoder~\cite{visiontransformer,attentionisallyouneed}, and harvest the respective features $V\in \mathcal{R}^{P\times D}$, $L\in \mathcal{R}^{\mathbb{L}_T\times D}$, and $E\in \mathcal{R}^{\mathbb{L}_E\times D}$, where $P$ is the number of image patches, $\mathbb{L}_T$ and $\mathbb{L}_E$ are the length of the query and explanation texts, respectively. For notational brevity, we reuse the symbols to represent their features.

\noindent\textbf{MMAE} takes the explanation text as auxiliary clues to narrow the semantic gap. In particular, MMAE comprises three groups of experts: image, query text, and explanation text groups. Each expert group is formed by several expert networks for feature encoding. Suppose there are $K$ experts in image expert group, each expert comprises several attention-based blocks. The image $V$ passes through every expert in parallel, we take the produced \texttt{[cls]} token feature ($\in \mathcal{R}^{D}$) as the output of the expert. Consequently, 
$K$ image features are obtained, marked as $\{\bar{V}_1,...,\bar{V}_K \}$. Following similar steps, we can acquire $M$ query text features $\{\bar{T}_1,...,\bar{T}_M\}$ from $M$ query text experts. 

Unlike the image and query text experts simply encoding the features, the explanation experts are responsible to allow the image and text to query clues from the explanation text, thereby narrowing the semantic gap. As shown in {Figure~\ref{framework}}, each explanation expert takes the image/the query text and the explanation text as input and outputs a bridge vector. To produce the image-oriented bridge vector, we take the image as query and perform a cross-attention procedure in each explanation expert:
\begin{align}
\label{eq1}
    \text{Attn}(V,E) &= \text{softmax}(\frac{W_qV\times W_kE}{\sqrt{D}})\times W_vE \\
    \label{eq2}
    \bar{E}^v &= (V + \text{MLP}(\text{LN}(\text{Attn}(V,E))))_{\texttt{cls}}
\end{align}
where $\bar{E}^v \in \mathcal{R}^D$ is the \texttt{[cls]} token vector, subscript $(\cdot)_{\texttt{cls}}$ means indexing the features of \texttt{[cls]} token. $W_k\in \mathcal{R}^{D\times D}$ is a linear projection matrix, $\text{LN}(\cdot)$ means the layer normalization, and $\text{MLP}(\cdot)$ indicates a two-layer linear projection with scale ration of 4 with GELU activation.  

Assume there are $N$ experts in explanation expert group, then we can obtain $N$ bridge vectors by feeding the image and the explanation text into each expert: $[\bar{E}^v_1,...,\bar{E}^v_N]$. Finally, we combine the vision features and visual bridge features to give a comprehensive representation for the image: 
\begin{equation}
    V_F = \{\bar{V}_1,\bar{V}_2,...,\bar{V}_K, \bar{E}^v_1,\bar{E}^v_2,...,\bar{E}^v_N\}.
\end{equation}
For the text comprehensive representation, we can acquire in a similar fashion by following Eq.\ref{eq1}-\ref{eq2}, marked as:
\begin{equation}
  T_F = \{\bar{T}_1,\bar{T}_2,...,\bar{T}_M, \bar{E}^t_1,\bar{E}^t_2,...,\bar{E}^t_N\}.   
\end{equation}

\noindent\textbf{Feature Aggregation.}
Subsequently, image/text features and queried bridge vectors are integrated to form an comprehensive representation. To well study the effectiveness of the acquired representations from MMAE, we design three different strategies of aggregation.

\noindent\textbf{\emph{Average}.} The average of all vectors is a na\"ive strategy without introducing any parameters. For image and text, their final representation can be represented as follows:
\begin{align}
    V^{*} = \frac{1}{|V_F|}\sum_{v\in V_F}v, \quad T^{*} = \frac{1}{|T_F|}\sum_{t\in T_F}t.
\end{align}

\noindent\textbf{\emph{Attention-based Aggregation}.} Treating $V_F$ and $T_F$ as feature sequences, we can aggregate them via the effective attention. Particularly, we first expand $V_F$ and $T_F$ with a vision token $\bar{V}_\texttt{vis} \in \mathcal{R}^{D}$ and text token $\bar{T}_\texttt{txt} \in \mathcal{R}^{D}$:
\begin{align}
    V^{'}_F = [\bar{V}_\texttt{vis},\bar{V}_1,\bar{V}_2,...,\bar{V}_K, \bar{E}^v_1,\bar{E}^v_2,...,\bar{E}^v_N], \\
    T^{'}_F = [\bar{T}_\texttt{txt},\bar{T}_1,\bar{T}_2,...,\bar{T}_M, \bar{E}^t_1,\bar{E}^t_2,...,\bar{E}^t_N].
\end{align}
$[\cdot,\cdot]$ means concate operation. We next perform an attention-based procedure:
\begin{align}
    V^{*} &= (V^{'}_F + \text{MLP}(\text{LN}(\text{Attn}(V^{'}_F,V^{'}_F))))_{\texttt{vis}} \\
    T^{*} &= (T^{'}_F + \text{MLP}(\text{LN}(\text{Attn}(T^{'}_F,T^{'}_F))))_{\texttt{txt}}
\end{align}

\noindent\textbf{\emph{Adaptive Gated Aggregation}} provides an adaptive integration strategy to aggregate the features.  First, the adaptive weights for image and text aggregation are generated as follows:
\begin{align}
    \mathcal{W}_v = \text{softmax}(V_{\texttt{cls}}W_v), 
    \mathcal{W}_t = \text{softmax}(T_{\texttt{cls}}W_t),
\end{align}
where $W_v\in \mathcal{R}^{D\times (K+N)}$, $W_t\in \mathcal{R}^{D\times (M+N)}$ are the linear projection matrix for image and text weights generation, and the $\mathcal{W}_v\in \mathcal{R}^{K+N}$, $\mathcal{W}_t\in \mathcal{R}^{M+N}$ are the generated adaptive weights for image and text.
Then, the enhanced image and text features are produced via a weighted integration:
\begin{equation}
    V^* = \sum_{i=1}^{K+N}\mathcal{W}_v[i]\cdot V_F[i],\quad T^* = \sum_{i=1}^{M+N}\mathcal{W}_t[i]\cdot T_F[i].
\end{equation}

With the enhanced features $V^*$ and $T^*$ from any of above integration strategies, we next impose a contrastive learning on them to optimize the network:
\begin{equation}
\label{itc}
\begin{split}
    \mathcal{L}_{VTC}(V^*, T^*) &= -\frac{1}{2}(\log\frac{\exp(s(V^{*},T^{*}))}{\sum_{T'\in \mathcal{B}}\exp(s(V^{*},T'))}\\
                 &+\log\frac{\exp(s(V^{*},T^{*}))}{\sum_{V'\in \mathcal{B}}\exp(s(V',T^*))}),
\end{split}
\end{equation}
where $\mathcal{B}$ is the training batch, $s(\cdot,\cdot)$ means the cosine similarity.

\begin{table*}[t]
\centering
\caption{\textbf{Quantitative results on VisualNews}. Three comparison groups from top to bottom subsequently shows the zero-shot performance, fine-tuned performance, and the variants of our EntityCLIP. \textbf{Bold} indicates the best perforamnce.
}
\vspace{-0.4cm}
\setlength{\tabcolsep}{2mm}
\setlength{\extrarowheight}{-4pt}
{
\renewcommand\arraystretch{0.65}\begin{tabular}{@{}lccccccccccc@{}}
\toprule[1.5pt]
\rowcolor[gray]{.95}\multirow{2}{*}{Models}  &\multicolumn{3}{c}{Image Retrieval} &\multirow{2}{*}{AVG t2i$\uparrow$} &\multirow{2}{*}{MR t2i$\downarrow$} & \multicolumn{3}{c}{Text Retrieval} &\multirow{2}{*}{AVG i2t$\uparrow$} &\multirow{2}{*}{MR i2t$\downarrow$}  \\ 
\cmidrule(lr){2-4} \cmidrule(lr){7-9}
\rowcolor[gray]{.95}& $K=1$ & $K=5$ & $K=10$  & &  & $K=1$ & $K=5$ & $K=10$  & &\\

\midrule
XFM~\cite{xfm} & 4.85 &12.28 &16.89	 &21.23	 & 2231.77 &5.89 &12.58	&16.79 &20.78	 & 1972.83	 \\
X-VLM~\cite{xvlm}  & 5.45 &13.04 &17.59 & 21.233 & 2530.78	& 6.29	&14.35	&19.31  &23.00 &1932.04 \\
X2-VLM~\cite{x2vlm}  & 11.57 &25.32 &32.71 & 36.40 & 1095.53 & 12.52	&26.32	&33.77 & 37.53 &865.47 \\
ALBEF~\cite{albef}  & 12.51	&25.09 &31.43 &34.18  &1830.19  &13.16	&26.22	&32.61 & 35.55  &1140.32 \\
EDIS~\cite{edis}  & 20.81	&38.21	&46.31 & 47.84 & 956.62 &  22.36	&40.77	&49.10 & 54.33 & 782.52 \\
BLIP~\cite{blip}  & 30.31 	&50.85	&58.90 & 59.83 & 133.23 &30.83	&51.72	&60.10 & 60.98 & 100.59 \\
BLIP2~\cite{blip2}  & 20.24	&37.57	&45.24 & 46.86 & 980.17 & 19.11	&36.64	&44.84 & 46.67 & 966.29 \\
$\text{CLIP}_{\text{ViT-B/32}}$~\cite{clip}  & 37.13	&59.68	&67.34 & 66.01 & 401.64 & 36.56	&58.16	&65.99 & 65.12 & 280.76 \\
$\text{CLIP}_{\text{ViT-B/16}}$~\cite{clip}  & 42.84 &65.06 &72.34 &	70.45 &345.28	&42.18	&64.02	&71.13 & 69.63 & 234.91 \\
\midrule

ALBEF~\cite{albef}  & 28.33	&49.27 &57.27 & 53.35  &821.30  &31.69	&52.44	&60.41 & 56.12  & 621.71  \\
BLIP~\cite{blip} & 34.20 	&55.23	&66.28 & 61.82 & 340.13 & 35.09	& 59.37	& 68.81 & 67.90 & 231.87 \\
$\text{CLIP}_{\text{ViT-B/32}}$~\cite{clip}   & 38.85	&62.95	&70.85	 &69.16	&226.88  & 40.138	&63.44	&71.21  & 69.72 & 125.83\\
$\text{CLIP}_{\text{ViT-B/16}}$~\cite{clip} & 42.06	&65.78	&73.59 &71.51 &182.81 &43.21	&65.96	&73.68 &72.02  &95.60 \\
\midrule
\textcolor{white}{$\text{EntityCLIP}_{\text{ViT-B/32}}$}\qquad AVG & 42.41	&65.93	&73.47 & 71.50  & 202.26	&43.82	&66.80	&74.14 & 72.45 &103.30 \\     
$\text{EntityCLIP}_{\text{ViT-B/32}}$\qquad Attn &42.26	&\textbf{66.52}	& \textbf{73.78} & 71.58  & 199.87	&43.80	&66.90	&74.22 & 72.51 &101.79 \\
\textcolor{white}{$\text{EntityCLIP}_{\text{ViT-B/32}}$}\qquad AGA & \textbf{42.91}	&66.09	&73.56 & \textbf{72.05} & \textbf{198.52} & \textbf{44.07} & \textbf{67.13} & \textbf{74.56} & \textbf{72.87} & \textbf{100.78}  \\
\textcolor{white}{$\text{EntityCLIP}_{\text{ViT-B/16}}$}\qquad  AVG &48.26	&71.23	&78.13 &75.54 &164.58 & 48.26	&71.23	&78.13 &76.59 &81.87 \\     
$\text{EntityCLIP}_{\text{ViT-B/16}}$\qquad Attn &  48.25 &\textbf{71.69}	&78.06 &75.61 &162.15 &\textbf{49.96} &\textbf{72.92} &78.99 &76.61 &79.14 \\
\textcolor{white}{$\text{EntityCLIP}_{\text{ViT-B/16}}$}\qquad AGA & \textbf{48.72}	&71.41  & \textbf{78.90} & \textbf{75.95} & \textbf{160.50} & 49.87 &72.18 	& \textbf{79.29} & \textbf{76.78} & \textbf{77.47} \\	
\bottomrule[1.5pt]
\end{tabular}}
\label{visualnews}
\end{table*}

\subsection{Gated Integrative Image-text Matching}
MMAE-derived features offer a holistic depiction of image-text pair. To further leverage these representations, we consolidate them and feed the result into image-text matching head to impose another constrain, thereby further facilitating the cross-modal alignment. 
GI-ITM initially consolidates the image-text representations into an unified multimodal representation via an adaptive aggregation mechanism, which is fed forward an image-text match head to give the matching probability of the image-text pair. In detail, the procedure can be formulated as:
\begin{equation}
\left\{\begin{aligned}
    &p(V, T) = \text{sigmoid}(\text{FC}(F_{mm}))\\
    &F_{mm} = \mathcal{W}_{mm} [V_F,T_F]\\
    &\mathcal{W}_{mm} =\text{softmax}([V_\texttt{cls},T_\texttt{cls}]W_{mm}),
\end{aligned}
\right.
\end{equation}
where 
$\mathcal{W}_{mm}\in\mathcal{R}^{K+M+2N}$, $W_{mm}\in\mathcal{R}^{2D\times (K+M+2N)}$, $p$ indicates whether input image-text is paired or not.

\begin{table*}[t]
\centering
\caption{\textbf{Quantitative results on GoodNews}. Three comparison groups from top to bottom subsequently shows the zero-shot performance, fine-tuned performance, and the variants of our EntityCLIP. \textbf{Bold} indicates the best perforamnce.
}
\vspace{-0.4cm}
\setlength{\tabcolsep}{2mm}
\setlength{\extrarowheight}{-10pt}
{
\renewcommand\arraystretch{0.65}\begin{tabular}{@{}lccccccccccc@{}}
\toprule[1.5pt]
\rowcolor[gray]{.95}\multirow{2}{*}{Models}  &\multicolumn{3}{c}{Image Retrieval} &\multirow{2}{*}{AVG t2i$\uparrow$} &\multirow{2}{*}{MR t2i$\downarrow$} & \multicolumn{3}{c}{Text Retrieval} &\multirow{2}{*}{AVG i2t$\uparrow$} &\multirow{2}{*}{MR i2t$\downarrow$}  \\ 
\cmidrule(lr){2-4} \cmidrule(lr){7-9}
\rowcolor[gray]{.95}& $K=1$ & $K=5$ & $K=10$  & &  & $K=1$ & $K=5$ & $K=10$  & &\\

\midrule
XFM~\cite{xfm} &1.95 &5.59	&8.42 & 12.29 &3340.41 &2.08 &5.80	&8.49 &12.21 &3191.23	 \\
X-VLM~\cite{xvlm}  & 3.32 &8.14	 &11.27	 &14.68	 & 3778.50 &3.98 &9.87	&13.57 &16.93	 & 2921.79 \\
X2-VLM~\cite{x2vlm}  & 3.95	&10.04 &13.96 & 18.03 &2694.95 & 4.20	&9.77	&13.17 & 16.39 &2807.28 \\
ALBEF~\cite{albef}  & 9.26	&19.78	&25.40 &28.82  &1930.8 &9.73	&20.30	&26.23 & 29.67 &1491.49 \\
EDIS~\cite{edis}  & 18.19	&33.75	&40.85 & 43.05 &1000.24 &19.85	&37.44	&45.28 & 47.12 &819.56\\
BLIP~\cite{blip}  & 15.05 &29.62 &36.80 & 39.73 & 1069.68 &15.72 &30.86	&38.48 & 41.36 &809.05 \\
BLIP2~\cite{blip2}  & 14.91	&28.90 &35.38 & 37.66 & 1614.76 & 13.93	&28.19 &35.09 & 38.05 & 1087.94 \\
$\text{CLIP}_{\text{ViT-B/32}}$~\cite{clip}  & 31.35	&53.24	&61.36 & 61.12 & 324.04 & 31.11	&52.21	&60.14 & 60.29 & 274.64 \\
$\text{CLIP}_{\text{ViT-B/16}}$~\cite{clip}  & 37.52	&60.52	&68.20 &	66.90 &253.83	& 37.49	&59.61	&67.05 & 66.25 & 200.12 \\
\midrule

ALBEF~\cite{albef}  &24.30	&50.29	&62.24  &59.07 &379.12  &22.72	&48.28	&60.33 &61.35 &249.99 \\
BLIP~\cite{blip} & 27.35 & 52.30 & 64.00 & 60.23 &211.90  & 30.32 & 53.34	& 64.31 & 61.31 & 198.45 \\
$\text{CLIP}_{\text{ViT-B/32}}$~\cite{clip}   & 34.77	&58.46	&67.09	 &66.24	&164.29  & 36.05 &59.60	&67.82  & 67.07 & 120.65\\
$\text{CLIP}_{\text{ViT-B/16}}$~\cite{clip} & 37.85	&62.22	&70.66 &71.51 &144.03 &39.36 &63.22	&71.05  & 69.88 & 98.45 \\
\midrule
\textcolor{white}{$\text{EntityCLIP}_{\text{ViT-B/32}}$}\qquad AVG & 36.13	&60.17	&68.48 & 67.38  & {160.87}	&37.94	&61.35	&69.38 & 68.41 &116.14 \\     
$\text{EntityCLIP}_{\text{ViT-B/32}}$\qquad Attn &36.14	&60.14	&68.59 & 67.39  & 161.07	&37.94	&61.34	&69.29 & 68.36 &115.75 \\
\textcolor{white}{$\text{EntityCLIP}_{\text{ViT-B/32}}$}\qquad AGA & \textbf{36.25}	& \textbf{60.24} &\textbf{68.68} & \textbf{67.48} & \textbf{159.47} & \textbf{38.03}	& \textbf{61.36} & \textbf{69.57} & \textbf{68.47} & \textbf{113.92}  \\
\textcolor{white}{$\text{EntityCLIP}_{\text{ViT-B/16}}$}\qquad  AVG & 41.48	&65.93	&73.62 &71.77 & 125.35 &43.50 &67.14 &74.37 &72.82 &83.74  \\     
$\text{EntityCLIP}_{\text{ViT-B/16}}$\qquad Attn &  42.79 &\textbf{67.24}	&74.71 &\textbf{72.61} &\textbf{123.79} &{44.90} & {68.18}	& {75.50} & {73.65} & {83.19} \\
\textcolor{white}{$\text{EntityCLIP}_{\text{ViT-B/16}}$}\qquad AGA & \textbf{42.90}	& {67.14} & \textbf{74.75} & {72.06} & {126.14} & \textbf{44.97} & \textbf{68.29} & \textbf{75.52} & \textbf{73.73} & \textbf{81.49}\\	
\bottomrule[1.5pt]
\end{tabular}}
\label{goodnews}
\end{table*}

The matched image-query text pairs in training dataset are taken as the positive pairs. For the negative pairs, we follow ALBEF~\cite{albef} to sample the negative samples. Particularly, for image $I$, a negative text is sampled with probability $\widetilde{T}\sim\mathcal{P}(\text{softmax}([s(V,T_1),...,s(V,T_{|\mathcal{B}|})]))$.  In analogy, the negative image $\widetilde{V}$  for query text $T$ can be picked. Then, the GI-ITM loss can be calculated:
\begin{equation}
\begin{split}
   \mathcal{L}_{GFM} &= \frac{1}{3}(-\log p(V,T) + (1-\log p(V, \widetilde{T})) \\
   &+ (1-\log p(\widetilde{V},T))).
\end{split}
\end{equation}

\begin{table*}[t]
\caption{\textbf{Comparison of Zero-shot Evaluation}. The trained on two large datasets, GoodNews or VisualNews, are directly evaluated on the other datasets to test the generality, where $IR$ and $TR$ refer to ``Image Retrieval'' and ``Text Retrieval'', CLIP* means CLIP trained on the source dataset, AVG = (AVG t2i + AVG i2t) / 2.
}
\vspace{-0.2cm}
\centering
\subfloat[
Trained on GoodNews and Evaluated on VisualNews and N24News.
\label{tab:ablation:pure_attentive}
]{
\centering
\begin{minipage}{0.92\linewidth}{\begin{center}
\setlength{\tabcolsep}{2mm}
\setlength{\extrarowheight}{-4pt}
{
\renewcommand\arraystretch{0.65}\begin{tabular}{@{}llccccccccccc@{}}
\toprule[1.5pt]
\rowcolor[gray]{.95}\multirow{2}{*}{Backbone} &\multirow{2}{*}{Models}  &\multicolumn{4}{c}{GoodNews$\rightarrow$ VisualNews}  &\multirow{2}{*}{AVG} & \multicolumn{4}{c}{GoodNews$\rightarrow$ N24News}  &\multirow{2}{*}{AVG}  \\ 
\cmidrule(lr){3-6} \cmidrule(lr){8-11}
\rowcolor[gray]{.95} & & $IR@1$ & $IR@5$ & $TR@1$  & $TR@5$ &  & $IR@1$ & $IR@5$ & $TR@1$  & $TR@5$ &\\
\midrule
\multirow{3}{*}{ViT-B/32} & CLIP~\cite{clip}  &37.45	& 59.68	&36.56  & 58.16 & 65.57 & 48.27	& 69.95	& 43.27 & 64.42 & 73.37 \\
&CLIP*~\cite{clip}  & 32.17	& 54.29	&33.46  & 54.91 & 62.10 & 48.87	& 72.96	& 46.47 & 70.23& 77.09 \\
&\textcolor{black}{$\text{EntityCLIP}$} & \textbf{38.66}	& \textbf{ 61.24} & \textbf{41.05} & \textbf{62.79} & \textbf{68.28} & \textbf{51.49} & \textbf{74.28} & \textbf{50.93} & \textbf{73.22} & \textbf{81.49}\\	
\midrule
\multirow{3}{*}{ViT-B/16} &CLIP~\cite{clip} & 42.84 & 65.06 &  42.18 & 64.02 & 70.04 & 55.01 &75.74 & 49.33 & 69.92 & 77.53 \\
&CLIP*~\cite{clip}  &38.05  &60.33 &39.24 &61.76  &67.49 &53.98 &76.58 &53.64 &75.70 & 80.58 \\
&\textcolor{black}{$\text{EntityCLIP}$} & \textbf{43.29}	& \textbf{65.59} & \textbf{45.38} & \textbf{67.18} & \textbf{71.72} & \textbf{56.97} & \textbf{78.59} & \textbf{57.12} & \textbf{77.63} & \textbf{82.27}\\
\bottomrule[1.5pt]
\end{tabular}}
\end{center}}\end{minipage}
}
\\
\subfloat[
Trained on VisualNews and Evaluated on GoodNews and N24News.
\label{tab:ablation:share-key}
]{
\begin{minipage}{0.92\linewidth}{\begin{center}
\setlength{\tabcolsep}{2mm}
\setlength{\extrarowheight}{-4pt}
{\renewcommand\arraystretch{0.65}\begin{tabular}{@{}llccccccccccc@{}}
\toprule[1.5pt]
\rowcolor[gray]{.95}\multirow{2}{*}{Backbone} &\multirow{2}{*}{Models}  &\multicolumn{4}{c}{VisualNews$\rightarrow$ GoodNews}  &\multirow{2}{*}{AVG} & \multicolumn{4}{c}{VisualNews$\rightarrow$ N24News}  &\multirow{2}{*}{AVG}  \\ 
\cmidrule(lr){3-6} \cmidrule(lr){8-11}
\rowcolor[gray]{.95} & & $IR@1$ & $IR@5$ & $TR@1$  & $TR@5$ &  & $IR@1$ & $IR@5$ & $TR@1$  & $TR@5$ &\\
\midrule
\multirow{3}{*}{ViT-B/32} & CLIP~\cite{clip}  &31.35	& 53.24	&31.11  & 52.21 & 60.71 & \textbf{48.27}	& \textbf{69.95}	& 43.27 & 64.42 & 73.37 \\
&CLIP*~\cite{clip}  &27.63	& 48.88	&27.33  & {48.81}  & 57.82  & 42.36 & 65.19	& 40.68 & 62.70& 71.04 \\
&\textcolor{black}{$\text{EntityCLIP}$} & \textbf{31.62}	& \textbf{54.18} & \textbf{32.26} & \textbf{54.40} & \textbf{62.33} & {47.22} & {69.88} & \textbf{44.91} & \textbf{66.90} & \textbf{74.32}\\	
\midrule
\multirow{3}{*}{ViT-B/16} &CLIP~\cite{clip} & \textbf{37.52} & 60.52 & 37.49  & 59.61& 66.58 & \textbf{55.01} & \textbf{75.74} & 49.33 & 69.92 & 77.53 \\
&CLIP*~\cite{clip}  &31.64  & 54.19 & 32.59 & 54.87  & 62.47 &47.98 &79.41 &45.72 &68.16 & 70.16 \\
&\textcolor{black}{$\text{EntityCLIP}$} & \textbf{37.52}	& \textbf{61.15} & \textbf{38.88} & \textbf{61.57} & \textbf{67.97} & {54.54} & {75.28} & \textbf{52.97} & \textbf{73.68} & \textbf{79.28}\\
\bottomrule[1.5pt]
\end{tabular}}
\end{center}}\end{minipage}
}
\label{zero_shot}
\vspace{-.5em}
\end{table*}

\subsection{Training Objectives}
Besides the constrains from Eq.~\ref{itc} and the GI-ITM, we also include the image and text representations from image and text encoders and impose the contrastive learning. Overall, our final training objective is formed by three terms: 
\begin{equation}
\label{finaloss}
   \mathcal{L} = \mathcal{L}_{VTC}(V_{\texttt{cls}}, T_{\texttt{cls}}) + \eta\mathcal{L}_{GFM} + \lambda\mathcal{L}_{VTC}(V^*, T^*).
\end{equation}
where $\lambda,\eta$ are two trade-off hyper-parameters.

During inference, we only use the features from the image encoder and text encoder to estimate the similarity, which is the same as CLIP. In other terms, the auxiliary clues from LLMs is only adopted during training, thereby introducing no inference burden.

\begin{table}[t]
\centering
\caption{\textbf{Performance comparison with CNN backbones (AGA stratetgy}. Two groups of comparison from top to bottom subsequently shows the quantitative comparison with Resnet50 and Resnet 101 backbones~\cite{resnet}.}
\vspace{-0.4cm}
\footnotesize
\setlength{\tabcolsep}{0.15mm}
\setlength{\extrarowheight}{-2pt}
{\begin{tabular}{@{}lccccccccc@{}}
\toprule[1.5pt]
\rowcolor[gray]{.95}\multirow{2}{*}{Models} & \multirow{2}*{Backbone} &\multicolumn{2}{c}{Image Retrieval} &\multirow{2}{*}{AVG t2i} &\multirow{2}{*}{MR t2i} & \multicolumn{2}{c}{Text Retrieval} &\multirow{2}{*}{AVG i2t} &\multirow{2}{*}{MR i2t}  \\ 
\cmidrule(lr){3-4} \cmidrule(lr){7-8}
\rowcolor[gray]{.95}& & $R@1$ & $R@5$   & &  & $R@1$ & $R@5$   & &\\
\midrule
&\multicolumn{9}{c}{\underline{\textbf{\quad N24News \quad}}}   \\
CLIP-ZS & \multirow{3}{*}{Resnet50} &46.72	&67.91 &73.57 &66.18 &45.95 &66.02 &72.47 &49.64 \\
CLIP-FT &  &48.45	&72.22 &77.44 &28.78 &51.29	&73.32 &78.70 &21.40 \\
EntityCLIP &  &\textbf{50.14}	&\textbf{73.30} &\textbf{78.25} &\textbf{28.00} &\textbf{53.32} &\textbf{74.61} & \textbf{79.56} & \textbf{21.36} \\
\midrule
CLIP-ZS & \multirow{3}{*}{Resnet101} &49.95	&70.51 &75.95 &52.58 &47.62 &68.09 &73.94 & 48.91 \\
CLIP-FT &  &52.71 &74.61 &79.84 &25.39 &54.31 &76.63 &81.09 &15.92 \\
EntityCLIP &  &\textbf{53.67} &\textbf{75.54} & \textbf{80.16} & \textbf{23.21} & \textbf{54.88} & \textbf{76.96} & \textbf{81.27} & \textbf{14.79} \\
\midrule
&\multicolumn{9}{c}{\underline{\textbf{\quad VisualNews \quad}}}   \\
CLIP-ZS & \multirow{3}{*}{Resnet50} &34.81	&56.94 &63.77 &464.42 &34.53 &55.55 &60.51 &353.81\\
CLIP-FT &  &34.99 &58.14 &65.57 &255.29 &36.98	&59.22 &66.57 &148.53 \\
EntityCLIP &  &\textbf{38.10}	& \textbf{61.58} & \textbf{68.11} & \textbf{237.56} & \textbf{40.83} & \textbf{63.05} & \textbf{70.79} & \textbf{128.97} \\
\midrule
CLIP-ZS & \multirow{3}{*}{Resnet101} &37.44	&59.83 &66.12 &419.10 &36.70 &58.21 &65.11 &312.42 \\
CLIP-FT &  &40.27 &63.88 &69.97 &212.18 &41.90 &64.82 &71.02 & 115.37 \\
EntityCLIP &  &\textbf{41.99} & \textbf{65.51} & \textbf{71.19} & \textbf{209.31} & \textbf{43.83} & \textbf{66.33} & \textbf{72.27} & \textbf{107.36} \\
\midrule
&\multicolumn{9}{c}{\underline{\textbf{\quad GoodNews \quad}}}   \\
CLIP-ZS & \multirow{3}{*}{Resnet50} &29.90 &51.23 &59.30 &368.94 &30.23 &50.84 &58.99 &313.22 \\
CLIP-FT &  &29.98 &52.80 &61.79 &208.06 & 32.29 &54.90 &63.31 & 154.11 \\
EntityCLIP &  &\textbf{32.23} & \textbf{55.55} & \textbf{63.74} & \textbf{199.68} & \textbf{35.45} & \textbf{57.87} & \textbf{65.60} & \textbf{146.10} \\
\midrule
CLIP-ZS & \multirow{3}{*}{Resnet101} &32.33	&54.66 &61.97 &354.58 &32.10 &53.45 &61.15 &296.14 \\
CLIP-FT &  &35.19 &57.28 &65.08 &183.19 & 37.19 & 59.73 & 66.88 & 149.37 \\
EntityCLIP &  &\textbf{35.50} & \textbf{59.63} & \textbf{66.82} & \textbf{164.21} & \textbf{38.54} & \textbf{61.73} & \textbf{68.56} & \textbf{120.48} \\
\bottomrule[1.5pt]
\end{tabular}}
\label{cnncomparison}
\end{table}

\begin{table}[t]
\centering
\caption{\textbf{News classification accuracy on N24News},``ZS'' refers to the zero-shot performance, ``FT'' indicates the fine-tuned results. \underline{Underline} is the second-best results except the classification-focused model MMNet~\cite{n4news}.}
\vspace{-0.4cm}
\footnotesize
\setlength{\tabcolsep}{0.6mm}
\setlength{\extrarowheight}{-2pt}
{\renewcommand\arraystretch{0.85}\begin{tabular}{lccccccc}
\toprule[1.5pt]
\rowcolor[gray]{.95}\multirow{2}{*}{Models} &\multicolumn{3}{c}{ViT-B/32}   & \multicolumn{3}{c}{ViT-B/16} & \multirow{2}{*}{\emph{MMNet}} \\
\rowcolor[gray]{.95}\cmidrule(lr){2-4} \cmidrule(lr){5-7}
\rowcolor[gray]{.95}&
\multicolumn{1}{c}{ CLIP-ZS} & 
\multicolumn{1}{c}{ CLIP-FT} &
\multicolumn{1}{c}{ EntityCLIP} & 
\multicolumn{1}{c}{CLIP-ZS} & 
\multicolumn{1}{c}{ CLIP-FT} &
\multicolumn{1}{c}{EntityCLIP} & \\
\midrule
Image  & 
35.97& 39.14 & \underline{40.06} &
37.97 & 41.61 & \underline{43.00} & \emph{54.34}\\
Headline & 
23.65 & 34.39 & \underline{42.78} & 30.00 & 31.56& \underline{44.55} &\emph{71.98} \\
Im. \&  Hdl. & 35.76 & 42.65 & \underline{50.49} & 42.26 & 44.15& \underline{54.00} &\emph{79.41}\\
\bottomrule[1.5pt]
\end{tabular}}
\label{n4new_cls}
\end{table}

\section{Experiment}
\noindent\textbf{Experiment Setup.} We utilize CLIP as the backbone for its robust image-text matching capabilities, developing two EntityCLIP variants with CLIP ViT-B/32 and ViT-B/16, initialized with CLIP's pretrained parameters. The Mistral-7B model generates explanation text using our template. Explanation text and query share text encoder. Adhering to CLIP's settings, images are resized to $224\times 224$, and text length is capped at 77th word. By default, our model uses four experts for vision, explanation, and text, \emph{i.e.,} $K=M=N=4$, with loss coefficients $\lambda=\eta=0.1$. The network, implemented in PyTorch and optimized by Adam on 1 A6000 GPU, varies batch sizes due to GPU memory constraints: 192 for ViT-B/32 with a learning rate of $1e^{-5}$, and 96 for ViT-B/16 with a learning rate of $2e^{-6}$.

\begin{figure*}
    \centering
    \includegraphics[width=1\linewidth]{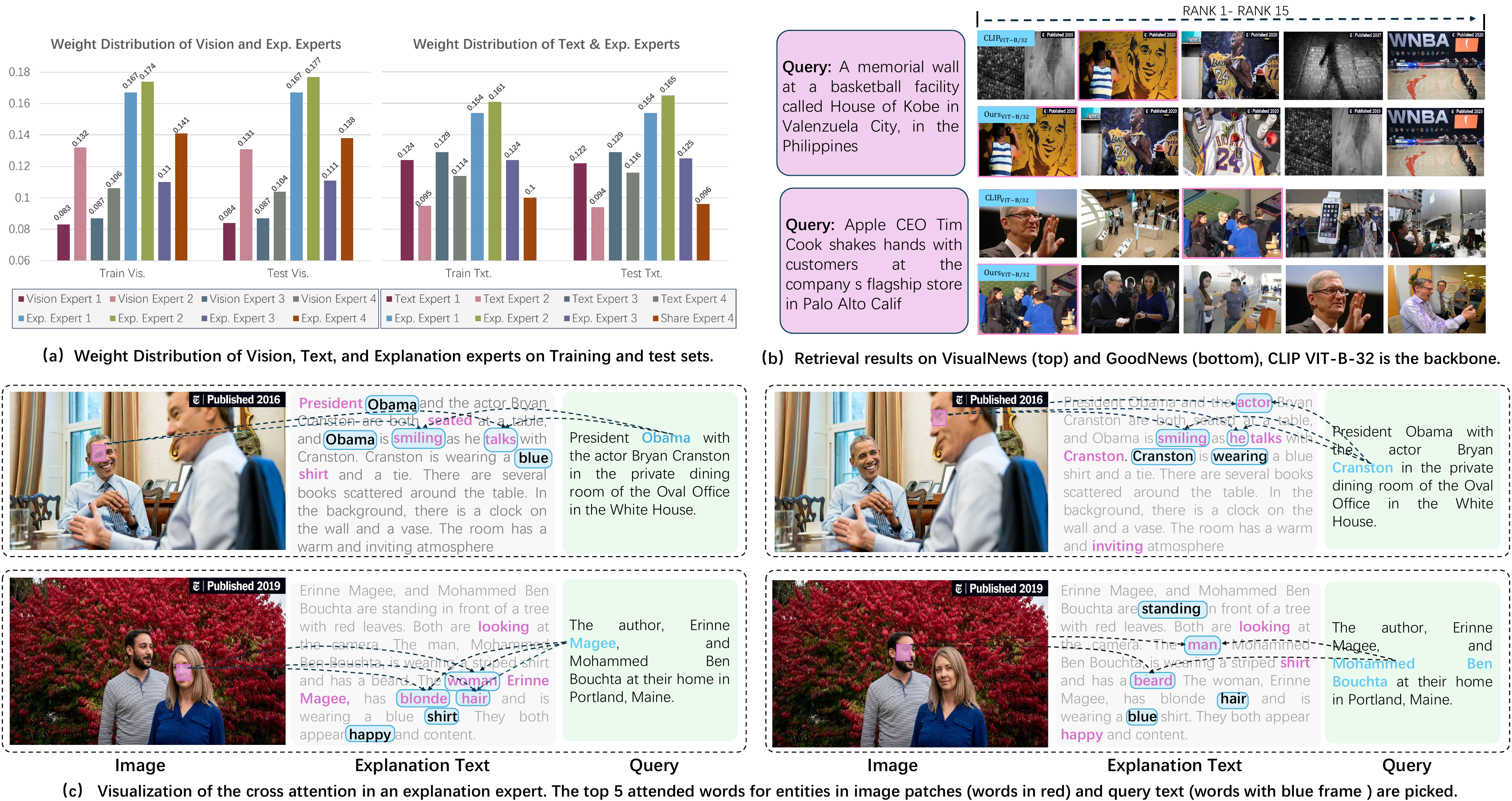}
    \vspace{-0.1cm}
    \caption{Visualization of the cross attention in an explanation expert. The top 5 attended words for entities in image patches (words in red) and query text (words with blue frame ) are picked.}
    \label{visresult}
\end{figure*}

\noindent\textbf{Datasets and Evaluation Metrics.} We adopt multimodal news dataset for evaluation, 
since the images in news associate with entity-related caption, which is suitable to evaluate entity-centric image retrieval. Three datasets are used for evaluation: N24News, VisualNews, and GoodNews. N24News~\cite{n4news}, sourced from The New York Times, contains 61,218 image-text pairs for multimodal news classification across 24 categories, with 48,988, 6,106, and 6,124 pairs allocated for training, validation, and testing. VisualNews~\cite{visualnews}, with 480,000 pairs from major news outlets, uses 400,000 for training, and 40,000 each for testing and validation. GoodNews~\cite{goodnews}, compiled from 2010 to 2018 New York Times articles, includes 488,986 image-entity caption pairs, utilizing 416,020 for training, and 24,205 and 48,761 for validation and testing. We employ standard recall metrics from the image-text matching community for evaluation, reporting both the average recall (AVG) on TOP 100 (=$(R@1+R@5+R@10+R@50+R@100)/5$) and the mean ranking (MR).

\begin{table*}[t]
\caption{\textbf{Ablation Experiments}. Sub-table (a)-(d) subsequently discuss the expert configuration in MMAE, the hyper-parameters $\eta$ and $\lambda$, the strategies of producing explanation text.
}
\vspace{-0.2cm}
\centering
\subfloat[
Expert Number Configurations in MMAE.
\label{tab:ablation:pure_attentive}
]{
\centering
\begin{minipage}{0.5\linewidth}{\begin{center}
\footnotesize
\setlength{\tabcolsep}{0.15mm}
\setlength{\extrarowheight}{-3pt}{
\renewcommand\arraystretch{0.85}\begin{tabular}{@{}lccccccccc@{}}
\multirow{2}{*}{Models} & \multirow{2}*{{\makecell*[c]{\#Experts\\ $\text{\fontsize{5.0pt}{\baselineskip}\selectfont[K,N,M]}$}}} &\multicolumn{2}{c}{Image Retrieval} &\multirow{2}{*}{\makecell*[c]{AVG t2i$\uparrow$}} &\multirow{2}{*}{\makecell*[c]{MR t2i$\downarrow$}} & \multicolumn{2}{c}{Text Retrieval} &\multirow{2}{*}{\makecell*[c]{AVG i2t$\uparrow$}} &\multirow{2}{*}{\makecell*[c]{MR i2t$\downarrow$}}  \\ 
\cmidrule(lr){3-4} \cmidrule(lr){7-8}
& & $R@1$ & $R@5$   & &  & $R@1$ & $R@5$   & &\\
\toprule[1.5pt]
&\multicolumn{9}{c}{\underline{\textbf{\quad N24News \quad}}} \\
Baseline & - & 51.52	 &73.87	 &78.84	&27.84  &50.90	&73.65  & 78.74 & 22.55\\
\makecell*[c]{Baseline\\ $\text{\fontsize{5.0pt}{\baselineskip}\selectfont+AVG E}$} & - &51.99	&64.22 &69.08 &265.21 &51.87 & 75.98 & 79.03 & 19.98 \\
\multirow{8}{*}{EntityCLIP} & [1,1,1] &53.54 &76.47 &80.39 &22.98 &54.77 & \textbf{76.55} &81.08 &15.73 \\
                      &[1,4,1] &53.27	&\textbf{76.68} &80.52 &\textbf{23.51} &54.69	&76.06 &80.89 &16.62 \\
                      &[2,2,2] &53.72	&76.37 & 80.46 &22.59 &54.88 &76.40 &81.05 & 16.41 \\
                      &[4,4,4] &\textbf{54.08} & {76.58} & \textbf{80.69} & {23.09} &\textbf{55.90} &{76.44} & \textbf{81.87} & \textbf{15.48} \\
                      &[4,8,4] &53.40	&75.88 &80.30 & 23.51 &54.07 &75.80 &80.66 & 16.53\\
\midrule
&\multicolumn{9}{c}{\underline{\textbf{\quad VisualNews \quad}}} \\
Baseline & - &39.83	&63.31 &68.97 &288.44 &40.43 &63.31 &69.19 &186.00 \\
\makecell*[c]{Baseline\\ $\text{\fontsize{5.0pt}{\baselineskip}\selectfont+AVG E}$} & - &40.72	&64.22 &69.08 &265.21 &41.82 &64.92 &70.88 &169.37 \\
\multirow{8}{*}{EntityCLIP} & [1,1,1] &41.99 &65.51 &71.19 &209.31 &43.83 &66.59 &72.31 &106.45 \\
                      &[1,4,1] &42.01 &65.34 &71.15 &210.15 &43.80	&66.53 &72.30 &107.62 \\
                      &[2,2,2] &42.14	&65.51 & 71.27 &212.99 &43.70 &66.47 &72.22 &106.15 \\
                      &[4,4,4] &\textbf{42.91}	&\textbf{66.52} & \textbf{72.05} & \textbf{198.52} & {44.07} & \textbf{67.13} & \textbf{72.87} & \textbf{100.78} \\
                      &[4,8,4] &41.31	&65.05 &70.80 &210.15 &43.09 &65.88 &71.81 &106.00\\
\midrule
&\multicolumn{9}{c}{\underline{\textbf{\quad GoodNews \quad}}} \\
Baseline & - & 34.77	&58.46	 &66.24	&164.29  & 36.05 &59.60  & 67.07 & 120.65\\
\makecell*[c]{Baseline\\ $\text{\fontsize{5.0pt}{\baselineskip}\selectfont+AVG E}$} & - &35.19	&58.92 & 66.81 & 163.82 & 36.88 & 60.07 & 67.88 & 118.91 \\
\multirow{8}{*}{EntityCLIP} & [1,1,1] &36.00	&59.89 &67.21 &162.17 &37.48 &60.75 &67.44 & 117.76 \\
                      &[1,4,1] &\textbf{36.29} &60.21 &67.42 & 159.83 & 37.97	&60.99 & 68.12 & 117.23 \\
                      &[2,2,2] & 36.18 &60.18 & 67.44 &161.47 & 38.00	& \textbf{61.47} & \textbf{68.47} &116.30 \\
                      &[4,4,4] &{36.25} & \textbf{60.24}  & \textbf{67.48} & \textbf{159.47} & \textbf{38.03}	& {61.36}  & \textbf{68.47} & \textbf{113.92}  \\
                      &[4,8,4] &35.39	&59.24 &66.82 & 162.55 & 37.24 &60.44 &67.82 & 116.51\\

\end{tabular}}
\end{center}}\end{minipage}
}
\subfloat[
Impact of Hyper-parameter $\eta$ in Eq~\ref{finaloss}.
\label{tab:ablation:share-key}
]{
\begin{minipage}{0.5\linewidth}{\begin{center}
\footnotesize
\setlength{\tabcolsep}{0.2mm}
\setlength{\extrarowheight}{0.1pt}{\renewcommand\arraystretch{1}\begin{tabular}{@{}lccccccccc@{}}
\multirow{2}{*}{Models} & \multirow{2}*{{\makecell*[c]{$\eta$}}} &\multicolumn{2}{c}{Image Retrieval} &\multirow{2}{*}{AVG t2i$\uparrow$} &\multirow{2}{*}{MR t2i$\downarrow$} & \multicolumn{2}{c}{Text Retrieval} &\multirow{2}{*}{AVG i2t$\uparrow$} &\multirow{2}{*}{MR i2t$\downarrow$}  \\ 
\cmidrule(lr){3-4} \cmidrule(lr){7-8}
& & $R@1$ & $R@5$   & &  & $R@1$ & $R@5$   & &\\
\toprule[1.5pt]
&\multicolumn{9}{c}{\underline{\textbf{\quad N24News \quad}}} \\
Baseline & - & 51.52	 &73.87	 &78.84	&27.84  &50.90	&73.65  & 78.74 & 22.55\\
\multirow{5}{*}{EntityCLIP} &0 & 52.22 & 74.98 & 79.72 & 34.01 & 53.19 & 75.27 & 80.31 & 18.44 \\ 
& 0.05 & {54.03} &\textbf{76.61} & 80.62& 23.31 & 55.82 & 76.44 & 81.21 & 16.49 \\
                      & 0.1 &\textbf{54.08} & {76.58} & \textbf{80.69} & \textbf{23.09} &\textbf{55.90} &\textbf{76.44} & \textbf{81.87} & \textbf{15.48} \\
                      &0.2 &54.03	&76.61 &80.51 & 23.55 &54.82 & {76.38} &81.26 &16.02 \\
                      &0.6 & 53.80 &75.82 & 80.04 &24.78 & 53.84	& 75.08 &80.13 & 17.12 \\
                      &1 &53.73 &75.66 & 79.94 & 24.99 &53.21 & 75.11 & 80.22 & 17.76 \\
\midrule
&\multicolumn{9}{c}{\underline{\textbf{\quad VisualNews \quad}}} \\
Baseline & - &39.83	&63.31 &68.97 &288.44 &40.43 &63.31 &69.19 &186.00 \\
\multirow{6}{*}{EntityCLIP} & 0 & 41.63 & 64.81 & 70.50 &215.18 & 42.76 & 66.59 &71.02 & 114.75 \\
                      & 0.05 & 42.12 & 65.27 &71.31 &207.12 & 43.59 & 66.62 &72.53 & 106.38 \\
                      & 0.1 &\textbf{42.91}	&\textbf{66.52} & \textbf{72.05} & \textbf{198.52} & \textbf{44.07} & \textbf{67.13} & \textbf{72.87} & \textbf{100.78} \\
                      &0.2 &42.82 &66.10 & 71.74 & 199.38 &43.76	&66.51 &72.64 &104.21 \\
                      &0.6 & 42.28 &65.87 &71.33 &203.56 & 43.41	&66.27 &72.21 & 106.92 \\
                      &1 &41.98	&65.86 & 71.63 &204.31 &43.56 & 66.52 & 72.41 &107.42 \\

\midrule
&\multicolumn{9}{c}{\underline{\textbf{\quad GoodNews \quad}}} \\
Baseline & - & 34.77	&58.46	 &66.24	&164.29  & 36.05 &59.60  & 67.07 & 120.65\\
\multirow{5}{*}{EntityCLIP} &0 & 35.29 & 59.76 &  67.03 & 163.01 & 37.11 & 60.98 & 67.87 & 117.83 \\
& 0.05 & 36.19 & {60.21} &{67.32} & {159.80} & {37.96} & \textbf{61.39} & \textbf{68.51} &116.62 \\
&0.1 & \textbf{36.25}	& {60.24}  & \textbf{67.48} & \textbf{159.47} & \textbf{38.03}	& {61.36}  & {68.47} & \textbf{113.92}  \\
&0.2 &36.21	&\textbf{60.27} & 67.39 &160.71 &37.88 &61.19 &68.44 &118.86 \\
&0.6 &35.82	&59.98 &66.90 &164.61 &37.08 &60.89 & 68.02 &118.97\\
&1 &35.85 &59.79 &67.02 &164.77 &37.14 &60.49 &67.81 &119.71 \\
\end{tabular}}
\end{center}}\end{minipage}
}
\\
\subfloat[
Impact of Hyper-parameter $\lambda$ in Eq~\ref{finaloss}.
\label{tab:ablation:share-key}
]{
\begin{minipage}{0.5\linewidth}{\begin{center}
\footnotesize
\setlength{\tabcolsep}{0.2mm}
\setlength{\extrarowheight}{-2pt}{\renewcommand\arraystretch{1}\begin{tabular}{@{}lccccccccc@{}}
\multirow{2}{*}{Models} & \multirow{2}*{{\makecell*[c]{$\lambda$}}} &\multicolumn{2}{c}{Image Retrieval} &\multirow{2}{*}{AVG t2i$\uparrow$} &\multirow{2}{*}{MR t2i$\downarrow$} & \multicolumn{2}{c}{Text Retrieval} &\multirow{2}{*}{AVG i2t$\uparrow$} &\multirow{2}{*}{MR i2t$\downarrow$}  \\ 
\cmidrule(lr){3-4} \cmidrule(lr){7-8}
& & $R@1$ & $R@5$   & &  & $R@1$ & $R@5$   & &\\
\toprule[1.5pt]
&\multicolumn{9}{c}{\underline{\textbf{\quad N24News \quad}}} \\
Baseline & - & 51.52	 &73.87	 &78.84	&27.84  &50.90	&73.65  & 78.74 & 22.55\\
\multirow{5}{*}{EntityCLIP} & 0.05 & \textbf{54.09} &76.44 &80.62 &208.31 & 54.80 & 76.42 & 81.04 & 16.53 \\
                      & 0.1 &{54.08} & \textbf{76.58} & \textbf{80.69} & \textbf{23.09} &\textbf{55.90} &76.44 & \textbf{81.87} & \textbf{15.48} \\
                      &0.2 &54.03	&76.44 &80.59 & 23.61 &54.61 &\textbf{76.49} &80.97 &16.58 \\
                      &0.6 & 53.67 &76.08 &80.32 &23.99 & 54.02	&75.90 &80.58 & 16.95 \\
                      &1 &53.69 &75.74 & 80.20 & 24.32 &53.71 & 75.51 & 80.35 & 17.38 \\
\midrule
&\multicolumn{9}{c}{\underline{\textbf{\quad VisualNews \quad}}} \\
Baseline   & - & 38.85	&62.95	 &69.16	&226.88  & 40.14	&63.44  & 69.72 & 125.83\\
\multirow{5}{*}{EntityCLIP} & 0.05  &42.05 &65.45 & 71.22 &208.31 & 43.66 &	66.59 &72.29 & 108.38 \\
 &0.1   &\textbf{42.91}	&\textbf{66.09}	 & \textbf{72.05} & \textbf{198.52} & \textbf{44.07} & \textbf{67.13}  & \textbf{72.87} & \textbf{100.78}  \\
 &0.2 &42.18	&65.81 &71.44 &200.83 & 43.50 &66.60 &72.28 &106.02 \\
 &0.6 &42.10 &65.64 &71.36 &203.84 & 43.50 &66.42 &72.18 &107.41 \\
 &1 & 42.01 &65.79 &71.33 &204.20 &43.57	&66.41 &72.17 &106.91\\
\midrule
&\multicolumn{9}{c}{\underline{\textbf{\quad GoodNews \quad}}} \\
Baseline & - & 34.77	&58.46	 &66.24	&144.03  & 36.05 &59.60  & 67.07 & 120.65\\
\multirow{5}{*}{EntityCLIP} & 0.05 & 36.22 & \textbf{60.33} &\textbf{67.49} & {159.80} & \textbf{38.06} & \textbf{61.43} & \textbf{68.55} &116.86 \\
&0.1 & \textbf{36.25}	& {60.24}  & {67.48} & \textbf{159.47} & {38.03}	& {61.36}  & {68.47} & \textbf{113.92}  \\
&0.2 &36.25	&60.22 &67.43 &160.89 &37.95 &61.29 &68.42 &119.01 \\
&0.6 &35.97	&60.05 &67.26 &163.57 &37.75 &60.95 &68.23 &118.67\\
&1 &35.88 &59.74 &67.12 &165.37 &37.62 &60.85 &68.05 &119.70 \\
\end{tabular}}
\end{center}}\end{minipage}
}
\subfloat[
Strategy Discussion of Producing the Explanation Text.
\label{tab:ablation:share-key}
]{
\begin{minipage}{0.5\linewidth}{\begin{center}
\footnotesize
\setlength{\tabcolsep}{0.05mm}
\setlength{\extrarowheight}{-2pt}{\renewcommand\arraystretch{1.1}\begin{tabular}{@{}lccccccccc@{}}
\multirow{2}{*}{Models} & \multirow{2}*{{\makecell*[c]{Explanation\\ Source}}} &\multicolumn{2}{c}{Image Retrieval} &\multirow{2}{*}{AVG t2i$\uparrow$} &\multirow{2}{*}{MR t2i$\downarrow$} & \multicolumn{2}{c}{Text Retrieval} &\multirow{2}{*}{AVG i2t$\uparrow$} &\multirow{2}{*}{MR i2t$\downarrow$}  \\ 
\cmidrule(lr){3-4} \cmidrule(lr){7-8}
& & $R@1$ & $R@5$   & &  & $R@1$ & $R@5$   & &\\
\toprule[1.5pt]
&\multicolumn{9}{c}{\underline{\textbf{\quad N24News \quad}}} \\
Baseline & - & 51.52	 &73.87	 &78.84	&27.84  &50.90	&73.65  & 78.74 & 22.55\\
\multirow{3}{*}{EntityCLIP} & Caption & 52.39 & 74.82 & 79.43 & 25.70 & 52.39 & 74.85 & 79.42 & 18.96 \\
                      & MLLM & 53.47 & 76.27 & 80.42 & 23.28 & 55.36 & 76.31 &80.61 & 15.82 \\
                      & LLM & \textbf{54.08} & \textbf{76.58} & \textbf{80.69} & {23.09} &\textbf{55.90} & \textbf{76.44} & \textbf{81.87} & \textbf{15.48} \\ 
                      \midrule
                      &\multicolumn{9}{c}{\underline{\textbf{\quad VisualNews \quad}}} \\
Baseline & - &39.83	&63.31 &68.97 &288.44 &40.43 &63.31 &69.19 &186.00 \\
\multirow{3}{*}{EntityCLIP} & Caption & 40.51 & 63.91 & 69.15 &264.32 & 41.28 & 64.47 &69.88 & 141.62 \\
                      & MLLM & 42.62 & 65.87 & \textbf{72.23} & 200.76 & 43.92 & 66.76 &72.07 & 104.36 \\
                      & LLM &\textbf{42.91}	&\textbf{66.52} & {72.05} & \textbf{198.52} & \textbf{44.07} & \textbf{67.13} & \textbf{72.87} & \textbf{100.78} \\
\midrule
&\multicolumn{9}{c}{\underline{\textbf{\quad GoodNews \quad}}} \\
Baseline & - & 34.77	&58.46	 &66.24	&164.29  & 36.05 &59.60  & 67.07 & 120.65\\
\multirow{3}{*}{EntityCLIP} & Caption & 35.12 & 58.93 &66.80 & 162.72 & 36.51 & 59.88 & 67.18 & 119.44 \\
                      & MLLM & 36.21 & 59.79 & 67.30 & 160.72 & 37.86 & 60.70 & 67.85 & 116.63 \\
                      & LLM & \textbf{36.25} & \textbf{60.24}  & \textbf{67.48} & \textbf{159.47} & \textbf{38.03}	& \textbf{61.36}  & \textbf{68.47} & \textbf{113.92}  \\ 

\end{tabular}}
\end{center}}\end{minipage}
}
\label{ablation}
\vspace{-.5em}
\end{table*}

\subsection{Quantitative Comparison}
We first report the zero-shot performance of large-scale pretrained multimodal models on the entity-centric retrieval datasets, \emph{to study whether the existing large-scale model can already address the entity-centric image retrieval}. We also report the fine-tuned performance of large multimodal models to make a fair comparison. 

\begin{figure}[t]
    \centering
    \includegraphics[width=1\linewidth]{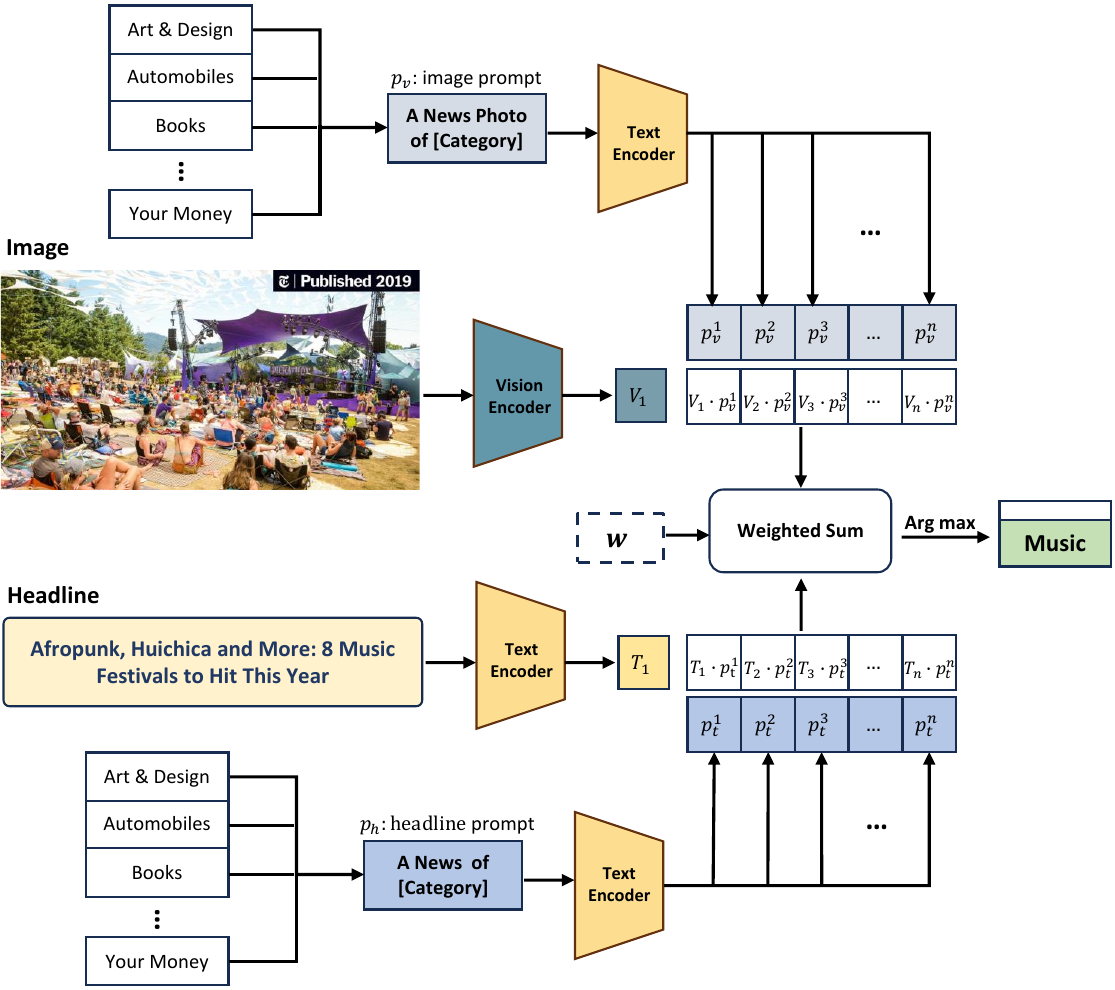}
    \caption{Illustration of utilizing the trained EntityCLIP without fine-tuning to perform multimodal news classification. The similarities from the image and the headline of the multimodal news are averaged as the final similarity score.}
    \label{news_cls}
\end{figure}

\begin{figure}
    \centering
    \includegraphics[width=1\linewidth]{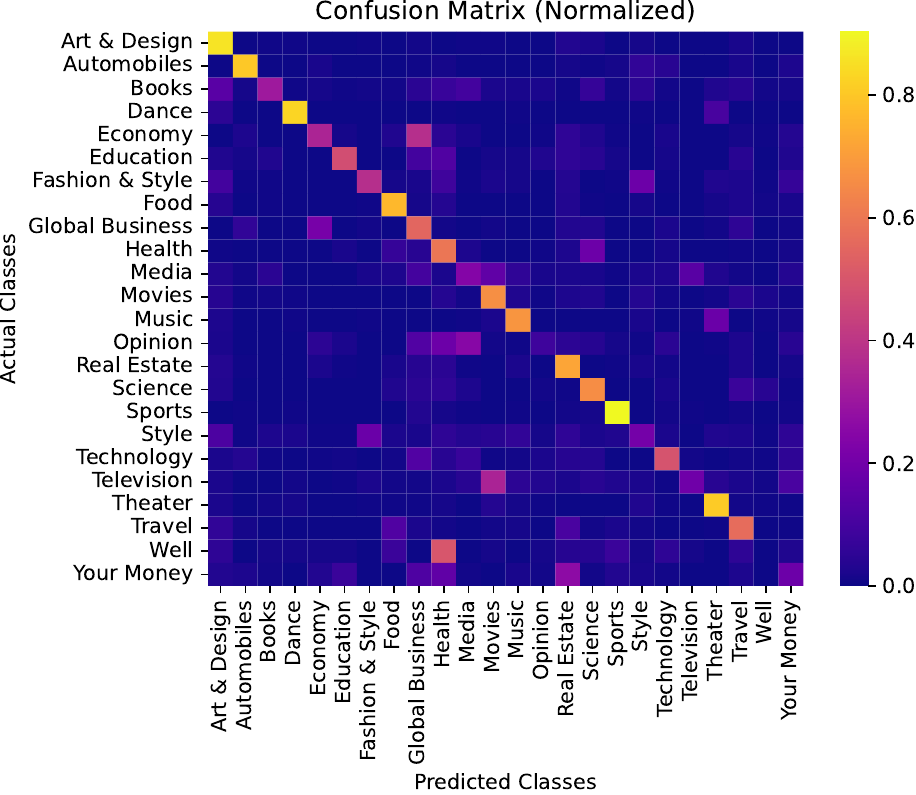}
    \vspace{-0.8cm}
    \caption{Confusion matrix visualization of multimodal news classification on N24News benchmark.}
    \label{conf_mat}
\end{figure}


The quantitative results across three datasets are detailed in Tables~\ref{n4news}-\ref{goodnews}. Initially, the first group for zero-shot performance comparison reveals that large-scale multimodal models fall short in addressing fine-grained entity-centric image retrieval. ALBEF, despite pretraining on 14M image-text pairs, achieves a Recall@1 of merely 21.18 on N24News and performs even less effectively on VisualNews and GoodNews. Similar limitations are observed with XFM, X-VLM, and BLIP. 
In contrast, CLIP demonstrates superior generalizability, due to its extensive (400M) dataset and contrastive learning approach, aligning more closely with retrieval objectives.

In the second comparison group, we report the performance of five methods fine-tuned on the entity-specific datasets. ALBEF shows a significant enhancement, with Recall@1 rising to 24.3 from 9.3 on GoodNews. However, CLIP's performance remains relatively stagnant, likely due to its training data's similarity to the News dataset, which may also explain its strong entity-centric query capability. Our method, leveraging CLIP as the base and employing various fusion techniques, markedly improves results. For instance, EntityCLIP with a ViT-B/32 backbone and a simple AVG fusion strategy achieves Recall@1 scores of 53.27, 42.41, and 36.13 on the N4News, VisualNews, and GoodNews datasets, respectively. Further, adaptive gate aggregation (AGA) boosts these figures, with our model outperforming the CLIP ViT-B/16 baseline by \textbf{6.7} in Recall@1 on the VisualNews dataset. Utilizing the ViT-B/16 backbone and AGA, EntityCLIP attains the highest scores: 60.85, 48.72, and 42.9 in Recall@1 for image retrieval across the datasets.

\noindent\textbf{Performance with CNN Backbones.} Our MMAE, designed with an attention-based architecture, is integrated with a vision transformer as the image encoder to optimize compatibility. To ascertain the efficacy of our module with CNN-based encoders, we conduct experiments using ResNet-50 and ResNet-101~\cite{resnet}. We adapt the $7 \times 7$ feature map from the CNNs to a sequence of 49 image tokens, employing the mean of these tokens as the \texttt{[cls]} token. This sequence is then processed by MMAE. The comparative results on three datasets, as detailed in Table~\ref{cnncomparison}, demonstrate that our approach consistently surpasses the performance of fine-tuned CLIP.

\noindent\textbf{Visualization.} We visually analyze the average weights of experts to assess their individual contributions in Figure~\ref{visresult} (a). We can observe two key points: (1) Training and testing sets exhibit similar weight distributions. (2) Explanation experts have higher weights than vision or text experts, underscoring the  
significant role of explanation text-derived features.  

In order to examine how image patches and textual words in query extract entity-related information from the accompanying explanation text, we conduct an attention visualization analysis on the explanation expert. Specifically, for a chosen explanation expert, we illustrate the top 5 attended words in the explanation text associated with entities in image regions and the query text. The results are depicted in Figure~\ref{visresult} (c), where the red words show the words attended by image patch, blue frames indicates the words attended by words in query.
It becomes evident that the entity-focused regions within the image and the entity terms in the query indeed concentrate on shared keywords. For instance, as seen in the upper-right subplot, both 'Cranston' patch in the image and the query word exhibit high attention towards the terms ``actor" and ``he", providing valuable cues for establishing connection between the visual and textual elements and bridging their semantic gap.

Figure~\ref{visresult} (b) illustrates a comparison of image retrieval results between the CLIP model and our EntityCLIP on VisualNews, and GoodNews datasets, presented sequentially from top to bottom. The figure demonstrates that our method outperforms CLIP for certain queries by relegating irrelevant images to lower ranks, thereby showcasing the enhanced discrimination capability of EntityCLIP in a visually intuitive manner.

\subsection{Generalization Evaluation.}

\noindent\textbf{Cross-Dataset Comparison.} Generalization ability is another important aspect for our EntityCLIP. To evaluate this, we perform a cross-dataset evaluation, \emph{i.e,}, test the models trained on one dataset using unseen datasets. The results are compared in Table~\ref{zero_shot}, where sub-table (a) reports the performance comparison  of model trained on GoodNews and tested on VisualNews and  N24News,  sub-table (b) is the comparison trained on VisualNews. We can observe that EntityCLIP outperforms the clip models in cross-dataset evaluations. Particularly, EntityCLIP trained on GoodNews shows better generality, surpassing CLIPs with a clear margin under all cases. The trained CLIP shows poor generalization ability, which means simply fine-tuning the models is not a robust solution for EITM problem.

\noindent\textbf{Zero-shot on Multimodal News Classification.}  We also extend our evaluation to Multimodal News Classification (MNC) on N24News dataset to further evaluate the generalization ability, employing prompts tailored for news images \( V \) and headlines \( H \): $P_v$ = \texttt{A News image of [News Category]}, $P_h$ = \texttt{A News of [News Category]}. 
Our experimental configuration closely mirrors that of CLIP, with a key distinction in the input modalities, which in our case are dual: the image and the headline within multimodal news, as depicted in Figure~\ref{news_cls}.
The association score \( A \) between the news and its category is calculated as \( A = w\cdot \text{sim}(V, P_v) + (1-w)\cdot \text{sim}(H, P_h) \), with \( w \) being a weighting parameter. Assigning news to the category with the highest association score, we set \( w = 1 \) and \( w = 0 \) to utilize only the image and headline, respectively. With equal consideration of both, we set \( w = 0.5 \). Table~\ref{n4new_cls} presents a comparison of classification accuracies; MMNet~\cite{n4news} is a model specialized for MNC, consequently, MMNet outperforms all EITM models in the table. Another important observation is that ETE's accuracy still significantly exceeds that of the fine-tuned CLIP, especially with ViT-B/16 backbone and multimodal inputs, EntityCLIP surpasses CLIP by nearly \textbf{10\%} accuracy.

As an exemplar of cross-modal retrieval models, EntityCLIP inherently encounters challenges in multimodal classification tasks due to its tendency towards semantic confusion among categories with similar semantics. This phenomenon is vividly illustrated in the confusion matrix depicted in Figure~\ref{conf_mat}. Specifically, the 'Television' category frequently gets misclassified as 'Movies', and the 'Economy' class is also prone to being conflated with the 'Global Business' category. These observation provide compelling evidence to explain why EntityCLIP demonstrates inferior performance compared to news classification-focused model MMNet. 

\subsection{Ablation Study}
We perform experiments on the three datasets using ViT-B/32 backbone to validate the effectiveness of our proposed components. The default configuration, as outlined in the experiment setup, serves as the basis for these analyses.

\noindent\textbf{Multimodal Attentive Experts.} 
Table~\ref{ablation} (a) evaluates performance across varying configurations of image (K), explanation (N), and text experts (M). We also study a na\"ive strategy to utilize the explanation text that averages the explanation text and the query features, denoted as Baseline + AVG E, which yields a modest improvement in Recall@1 to 39.8 on the image retrieval task. Introducing additional experts enhances performance; for instance, with K, N, M set to [2, 2, 2], Recall@1 increases to 42.14. The optimal configuration at K, N, M = 4 peaks at a Recall@1 of 42.91 and an average recall on TOP 100 of 72.05, significantly outperforming the baseline by 3.08 in average recall.

\noindent\textbf{Impact of hyper-parameter $\eta$ and $\lambda$.} The hyperparameters $\eta$ and $\lambda$ in our model are tasked with balancing the three constituent losses, with default values set at 0.1. This section delves into the effects of varying these parameters. 
Table~\ref{ablation} (b) examines the impact of $\eta$, varying from 0 to 1. The case of $\eta$=0 signifies the absence of GI-ITM, relying solely on MMAE, which still achieves a respectable Recall@1 of 41.63 on VisualNews, thereby highlighting the effectiveness of MMAE. The consistent outperformance of the baseline by all $\eta \neq 0$ configurations confirms the value of GI-ITM in conjunction with MMAE. The selection of $\eta=0.1$ propels our model to its peak performance. Parallel observations are drawn from the performance on N24News and GoodNews datasets.

Table~\ref{ablation} (c) presents a comparative analysis of EntityCLIP's performance across a range of $\lambda$ values from 0.05 to 1 on three datasets. The results indicate that each $\lambda$ value enhances performance to some extent; specifically, on VisualNews, $\lambda=0.05$ significantly exceeds the baseline. An optimal performance is achieved with $\lambda=0.1$, reaching a Recall@1 of 42.91 in image retrieval on VisualNews. Increments beyond this value do not yield further improvements, as evidenced in the table.

\noindent\textbf{Strategies of Producing Explanation Text}. We study three off-the-shelf models to produce explanation text: caption model BLIP2~\cite{blip2}, multimodal large language model Qwen-VL~\cite{qwen}, and large language models Mistral-7B.  Results in Table~\ref{ablation} (d) show that image captions slightly improve Recall@1 in image retrieval, edging out the baseline at 39.83 to 40.51. However, captions fall short in bridging the semantic gap compared to explanation texts, as evident in the table. This gap is attributed to captions typically providing only a general image description, lacking the specificity needed for entity-centric queries. In contrast, LLM-generated explanation texts establish connections with both images and query texts, as illustrated in Figure~\ref{exptxt}. While MLLM produces quality explanations, it requires more inference time and slightly underperforms compared to LLM.






\section{Conclusion}

This paper delves into the intricate challenge of fine-grained, entity-centric image retrieval, characterized by a significant semantic gap between entity-related texts and image contents. To surmount this obstacle, we harness the expansive knowledge base of LLM to narrow the semantic gap and develop a framework, termed EntityCLIP. EntityCLIP commences with extracting entity metadata from LLM, subsequently employing a meticulously crafted Multimodal Attentive Experts (MMAE) module to integrate the metadata for semantic gap narrowing. Further enhancing our methodology, we introduce a Gated Integrative Image-text Matching mechanism to utilize the rich features of the text-image pair, imposing image-text matching constraints. Our method's efficacy is substantiated through rigorous experimentation across three benchmark datasets.

\bibliographystyle{ACM-Reference-Format}
\bibliography{main}

\end{document}